%% file: main.tex
\documentclass[nonatbib]{article}
\usepackage{lipsum, lineno}
\usepackage{graphicx}
\usepackage{amsmath,amsfonts,amssymb}
\usepackage[preprint]{neurips_2020}
\usepackage[utf8]{inputenc} 
\usepackage[T1]{fontenc}    
\usepackage[breaklinks]{hyperref}
\usepackage[noabbrev]{cleveref}
\usepackage{url}            
\usepackage{booktabs}       
\usepackage{nicefrac}       
\usepackage{microtype}      
\usepackage{lipsum, lineno}
\usepackage{setspace}
\usepackage{cite}
\usepackage{algorithm}
\usepackage{algpseudocode}
\usepackage{tabularray}
\usepackage{multirow}
\usepackage{placeins}

\title{Sexual dimorphism and sex-specific features of pediatric hand bone shapes via statistical shape modeling}

\author{
  Edwin Tay$^{*}$, Amir A. Zadpoor$^{\dagger}$, Nazli Tümer$^{\dagger}$\\
  \\
  \textit{Department of Biomechanical Engineering, Faculty of Mechanical Engineering,}\\
  \textit{Delft University of Technology (TU Delft), Mekelweg 2, 2628CD, Delft, The Netherlands}\\
}

\begin{document}

\maketitle

\begin{abstract}

	Understanding the envelope of shape variation encompassing both sexual dimorphism and growth-related changes in living tissue is a non-trivial task. In this work, we use statistical shape models (SSMs) and analyze demographic-stratified shape parameters to quantify sexual dimorphism and sex-specific shape changes. To this end, we developed a two-stage detection-segmentation framework that accurately segmented $n = 27664$ unique bone masks from a radiographic dataset of $n = 1456$ subjects with diverse ages and sexes (HD95 = $1.512 \pm 0.005$, ASSD $0.449 \pm 001$, Dice = $0.987 \pm 0.00004$, IoU = $0.975 \pm 0.0001$). Following SSM fitting, our follow-up analyses of stratified shape parameters highlighted the morphology of sexual dimorphisms for each bone and demonstrated sex-specific differences in shape development. Our results were consistent with existing literature and provided an unambiguous, data-driven methodology for characterizing tissue shape and its population-scale variation. This approach provides a potential methodological foundation for developing population-based models of anatomical variation and could be extended to investigate tissue growth and shape changes over time. Future work should evaluate the applicability of this methodology on alternative anatomical structures and imaging modalities, with the goal of validating these growth trajectories on longitudinal data. 

\end{abstract}


\doublespacing

\input{mainmatter/1_introduction}


\input{mainmatter/2_methodology}


\input{mainmatter/3_results}


\input{mainmatter/4_discussion}


\input{mainmatter/5_conclusion}


\input{mainmatter/6_competing}

\input{mainmatter/8_acknowledgements}

\vfill

\pagebreak
\bibliographystyle{IEEEtran}
\bibliography{references.bib}
\vfill
\pagebreak

\input{mainmatter/9b_table_captions}

\input{mainmatter/9a_figure_captions}

\input{mainmatter/10_figures}

\input{mainmatter/7_supplementary}

\end{document}

%% file: mainmatter/1_introduction.tex
\section{Introduction}
	
	Growth and development are two defining characteristics of all living organisms. They involve a complex set of multifaceted processes that change an organism's form and function over its lifetime \cite{Goriely2017, POORTER2013, ballock2003growth}. These processes work in concert, an intricate orchestra of interdependent mechanisms influenced by internal (\textit{e.g.,} hormonal, genetic) and external (\textit{e.g.,} environmental) stimuli \cite{Gilbert2001, Arthur2002, Devesa2016}. These dynamic processes lead to distinct changes in anatomical structure configuration, providing morphological hallmarks of development over one's lifetime. Thus, adding temporal variation expands the shape space of potential anatomical variations. Characterizing this spatiotemporal space is, however, a complex proposition, particularly in population-level analyses, where inter-individual variability adds another layer of complexity. Nonetheless, uncovering the overarching envelope of population-level anatomical shape variation and how it evolves across developmental stages remains an open challenge.  
	
	In the context of growth and development, hand bones, particularly the phalanges and metacarpals, are a valuable focus of study. These bones are key components of upper-body mobility \cite{Maw2016}, along with their associated ligaments and muscles, which enable advanced manipulation capabilities such as prehension \cite{schwarz1955anatomy}. Therefore, injuries or pathological conditions, especially early in life, can severely impact overall mobility, further development, and quality of life and productivity \cite{Barton1979, dePutter2012, Chang2011}. Thus, understanding potential shape variations in these bones could inform pathological diagnosis and the design of medical interventions. Analyses of hand bone morphology are also utilized in forensic, archaeological, and anthropological contexts to predict bone and, by extension, chronological age \cite{Pinchi2014, Marzke1986, Kalichman2008}. To summarize, developing a holistic understanding of how living, growing hand bones change is essential for a variety of fields. While a range of studies characterize these changes, they do so through statistical analyses of osteometric measurements. These measurements, in turn, rely on exhaustive placement of consistent anatomical landmarks, which can prove unreliable and inconsistent \cite{Smith2013}. Thus, alternative methods should be extended and utilized, with their efficacy quantified. 

	Statistical shape models (SSMs) are statistical frameworks used to characterize the mean shape and the principal modes of variation within a population of shapes \cite{Sarkalkan2014}. In their seminal work, Cootes and Taylor introduced SSMs in the form of point distribution models (PDMs), which rely on sets of corresponding landmarks consistently defined across all shapes in a dataset \cite{Cootes1995}. After aligning the landmark configurations, a mean shape is computed, and principal component analysis (PCA) is applied to identify the principal modes of shape variation. Although alternative SSM media exist, such as interacting particles \cite{Cates2017}, level sets \cite{Kozic2008}, and diffeomorphisms \cite{Durrleman2014}, the PDM-based representation remains more intuitive and is therefore preferred \cite{Adams2022}. SSMs themselves have also been extended for applications such as partial shape reconstruction \cite{Zhang2016} and segmentation \cite{Heimann2009}. Regardless, the utility of SSMs lies in their capability to distill a dataset of shapes into a compact statistical model via a mean shape and principal modes of variation (\textit{i.e.,} shape modes) \cite{Paassen2024, Eijkenboom2023, Tmer2021}. Thus, investigating the applicability of SSMs to characterize and encompass the shape space of anatomical structures at different developmental time points is a compelling area of study. 
		
	Several studies have employed SSMs to characterize growth and morphological changes over time in a variety of anatomical structures. de Vries \textit{et al.} conducted a literature review of existing spatiotemporal shape modeling work for various anatomical structures \cite{deVries2025}. They observed correlations between the first principal mode of variation and age across anatomical structures, indicating that growth and related changes are an important dimension to consider. Shi \textit{et al.} further demonstrated that SSMs built with pediatric data are able to better reconstruct lower-limb pediatric bones from sparse inputs compared to those built on adult data \cite{Shi2022}. Munyebvu \textit{et al.} also developed an SSM of the proximal, medial, and distal phalanges, focusing on variability in the interphalangeal joints \cite{Munyebvu2024}. They demonstrated a unique methodology to neutralize unwanted joint angle variation during imaging, enabling an overarching analysis of the joints. Schneider \textit{et al.} also developed SSMs to compare and contrast differences in morphologies of male and female carpometacarpal joint bones \cite{Schneider2015}. They found that, discounting scaling-based differences, male and female bones are similarly shaped. Overall, while most of the aforementioned studies use 3D data (CT and MRI), the datasets are relatively small. Furthermore, they only investigate a limited scope of hand bones (\textit{i.e.,} 1 or 3 bones at most). Finally, the datasets are limited to very narrow age ranges, mostly near skeletally mature subjects (NSM). Thus, this significant gap in our understanding warrants further investigation. 
	
	In this study, we seek to systematically uncover and characterize the overarching envelope of shape variation across the bones of the fingers, that is, the metacarpals and proximal, medial, and distal phalanges. To do so, we used a PDM-based SSM framework in which we constructed separate SSMs for each bone type, collectively capturing variation across the finger bones. With the built SSMs, we can isolate and scrutinize subpopulations within their constituted SSM shape spaces to study sexual dimorphisms and shape changes over time in greater detail. To this end, we can garner additional data-driven insights into the overall shape envelope encompassing hand bone morphology and present a potential framework for use on other anatomical structures.

%% file: mainmatter/2_methodology.tex
\section{Methods}
	
	\subsection{Datasets and pre-processing}
    
	In this study, we utilized radiographs from the open-source RSNA Pediatric Bone Age Challenge (2017) dataset \cite{Halabi2019} (Figure 1A). We first pre-processed these data with contrast-limited adaptive histogram equalization (CLAHE) to enhance edges and improve segmentation outcomes \cite{Yoshimi2024}. From the overall RSNA dataset, we extracted $n_{raw} = 1456$ unique radiographs from individual subjects. We randomly selected these data, ensuring a relatively balanced sex distribution and a homogeneous age distribution (Table 1). 
	
	\subsection{Detection -- segmentation framework}

		To obtain segmentation masks for each finger bone in our dataset, we developed a two-stage detection-segmentation framework (Figure 1B). Herein, the framework first obtained labeled bounding boxes isolating each bone of interest. We then fed these bounding boxes into a universal bone segmentation head, which segments the bone within. To train this framework, we semi-automatically generated a dataset of images with bounding box labels from a semantic segmentation dataset. This dataset was a combination of manual annotations utilized in a previous study \cite{melol}, supplemented with additional segmentation masks generated by PedVision \cite{Homayounfar2024}. From an initial dataset of 583 images, we generated labels for 19 bones for each image, resulting in a total of 11077 unique bounding boxes. We then trained and evaluated alternative detection models utilizing an 80-10-10 training-validation-test split to determine the best configuration for our framework. Specifically, we utilized YOLOv8, YOLOv11, YOLOv12, and their varying model sizes to determine the best-performing model for our framework \cite{yolov8_ultralytics, yolo11_ultralytics, tian2025yolov12}. We utilized mAP50-95 as our principal evaluation metric for detector performance. In detail, this metric represents mean average precision averaged over intersection-over-union (IoU) thresholds of 0.5 to 0.95 in steps of 0.05. 

		For our universal bone segmentation head, we explored alternative model types and configurations. Specifically, we utilized DeepLabv3+ \cite{chenzhu18}, nnUNet \cite{Isensee2020}, SegFormer \cite{XieWang21}, Segment Anything 2 \cite{RaviGabeurHu24}, and MedSAM2 \cite{MaYangKim25} to compare convolution-based, transformer-based, and foundation-model-based segmentation heads. For all applicable architectures, we tested varying backbone sizes (\textit{e.g.,} MiT B0-B5 for SegFormer). Generally, for all explored models, we used frozen encoders and fine-tuned only the decoders on our image-mask dataset to leverage pre-trained weights. We utilized similar training-validation-test splits and evaluated segmentation outcomes via a ranking score based on the sum of average symmetric surface distance (ASSD) and 95th percentile Hausdorff distance (HD95). We also calculated Dice score and IoU, which were relatively saturated; however, we report them for completeness. We then used the best-performing detection and segmentation models to segment the extracted $n_{raw}$ radiographs, producing $n_{masks} = 27664$ unique bone masks.

	\subsection{Statistical shape modeling}
		
		In this study, we utilized the Python package \textit{pyssam} and its implementation of SSMs \cite{pyssam}. To build a PDM-based SSM, we first performed key preprocessing steps (Figure 1B). First, we aligned the data using an iterative closest points (ICP) algorithm \cite{Besl1992}. Then, we defined, propagated, and registered a consistent set of landmarks across each dataset using a coherent point drift (CPD) framework \cite{Myronenko2010}. This discretized each shape ($x_{i}$) in our dataset of ($i$) shapes and represented it via landmarks. For each shape, we translated the landmark configuration to its centroid and scaled it to unit centroid size, so the subsequent SSMs captured shape variation independent of translation and absolute size. Based on this correspondence, we a estimate a mean shape ($\bar{x}$) (Equation \ref{eq:mean}). 
			
			\begin{equation}
				\bar{x} = \frac{1}{N} \sum_{i=1}^{N} x_{i}
				\label{eq:mean}
			\end{equation} 
		
			A covariance matrix ($\mathbf{S}$) estimating the variation of each shape $x_{i}$ from $\bar{x}$ can also be calculated (Equation \ref{eq:covariance}).  
		
			\begin{equation}
				\mathbf{S} = \frac{1}{N} \sum_{i=1}^{N} (x_{i} - \bar{x}) (x_{i} - \bar{x})^\intercal
				\label{eq:covariance}
			\end{equation}
		
			Carrying out PCA on $\mathbf{S}$ results in eigenvectors ($\phi_{s}$) and their corresponding eigenvalues ($\lambda_{s}$). Thus, any shape ($x$),  whether within or outside the training dataset, can be described as follows (Equation \ref{eq:ssm}).
		
			\begin{equation}
				x = \bar{x} + \sum_{s=1}^{c} b_{s} \phi_{s}
				\label{eq:ssm}
			\end{equation}
	
			where $b_{s}$ is the shape parameter weighting the $s$\textsuperscript{th} mode of variation $\phi_{s}$, and the corresponding eigenvalue $\lambda_{s}$ gives the variance of $b_{s}$ across the training dataset. $c$ is the analysis threshold, \textit{i.e.,} the number of modes contributing most to shape variation. In this study, we set $c$ to the number of shape modes needed to account for $90\%$ of variation, ensuring sufficient anatomical variation is retained. Furthermore, while we grouped and analyzed all bone masks, we focused on the third ray for brevity. Nevertheless, we report results for the remaining bones in the supplementary material and briefly discuss them in the main text. For the remainder of this work, we refer to shape modes and modes of variation interchangeably and denote them as PC$\Phi$, where $\Phi$ is a suffix referring to the $s$th shape mode. 

	\subsection{Sexual dimorphisms}
		
		To quantitatively assess the prevalence of sexual dimorphism for each bone, we performed statistical analyses on the shape parameters extracted from each bone of each subject in our dataset. From our fitted SSMs, we obtained the shape parameters of $c$ shape modes (\textit{i.e.,} the analysis threshold) for each bone of each subject. We then stratified these data into sex-specific groups and utilized Hotelling's $T^2$ test, similar to the analysis of Moscoloni \textit{et al.} \cite{Moscoloni2025}. Briefly, Hotelling's $T^2$ is a multivariate extension of the Student's $t$-test that determines whether the mean vectors of several variables differ significantly between two populations. This test yields a $T^2$ statistic, which can be used to determine an $F$-value and corresponding $p$-value for significance testing. A Mahalanobis distance $\mathcal{D}$, related to $T^2$, can also be extracted. This quantifies the distance between the male and female centroids in the $c$-dimensional shape space. Thus, while the $p$-value indicates significance, $\mathcal{D}$ quantifies it independently of sample size. 
				
		We also extended this analysis to evaluate, in further detail, the sexual dimorphism of each shape mode. Welch's $t-$test was utilized to determine, on a per-mode basis, if there were statistically significant differences in the shape modes between sexes, which would indicate sexual dimorphism.  We corrected the $p$-values using the Benjamini-Hochberg procedure to determine significance, and we also extracted Cohen's $d$ to quantify separation between sex-stratified shape parameters.   
		
	\subsection{Association between sexual dimorphism and age}

		Age-related growth and development significantly affect skeletal morphology \cite{Goriely2017}. Nevertheless, to establish a deeper understanding of the relationship between sexual dimorphism and age, we developed our modeling approach further. We modeled each shape parameter $y_{i,j}$ of individual $i (i = 1, ..., n_{raw})$ for shape mode $j (j = 1, ..., c)$ as a multivariate general linear model with skeletal age, sex, and their interaction (Equation \ref{eq:mvglm}). The functional form of the age term was selected as a cubic based on initial findings (Appendix A). 

		\begin{equation}
			y_{i,j} = \beta_{0,j} + \sum_{k=1}^{3} \beta_{k,j} a_{i}^{k} + \gamma_{j} s_{i} + \sum_{k=1}^{3} \delta_{k,j} s_{i} a_{i}^{k} + \epsilon_{i,j}
			\label{eq:mvglm}
		\end{equation}

		where $a_{i}$ is skeletal age centered at cohort means and used instead of raw skeletal age to improve numerical stability in the presence of squared and cubed terms. $\gamma_{j}$ is the sex difference at the centering age, and $s_{i} \in \{0,1\}$ indicates female or male sex respectively. $\beta_{1,j} - \beta_{3,j}$ describes the female shape trajectory of mode $j$, whilst $\delta_{1,j} - \delta_{3,j}$ quantifies the departure of the male from the female trajectory. $\epsilon_{i,j}$ denotes residuals related to natural biological variability, that is, the variation not accounted for by age or sex. By collecting all $c$ shape modes, we can stack them to jointly model all $c$ modes and form the multivariate general linear model (Equation \ref{eq:mvglmstack}).
		
		\begin{equation}
			\mathbf{Y} = \mathbf{XB} + \mathbf{U}
			\label{eq:mvglmstack}
		\end{equation} 	 
	
		, where $\mathbf{Y} \in \mathbb{R}^{n_{raw} \times c}$, design matrix $\mathbf{X} \in \mathbb{R}^{n_{raw} \times q} (q = 8)$ whose $i$th row is $[1, a_{i}, a_{i}^{2}, a_{i}^{3}, s_{i}, s_{i}a_{i}, s_{i}a_{i}^{2}, s_{i}a_{i}^{3}]$, and $\mathbf{B} \in \mathbb{R}^{q x c}$ is estimated by ordinary least squares. $\mathbf{U}$ represents the residuals, which were assumed to be independent across individuals. The age $\times$ sex interaction was tested as the joint null hypothesis $\delta_{1} = \delta_{2} = \delta_{3} = 0$, that is, all interaction coefficients vanish, and both sexes follow identical trajectories separated by a constant offset. The test statistic was Pillai's trace, which summarizes the strength of association between the interaction terms and the shape modes after adjusting for the remaining terms in the model. Because the hypothesis comprises three interaction terms, Pillai's trace ranges from 0 to $s = \min(3, c) = 3$ rather than from 0 to 1. We therefore report both Pillai's trace and the normalized value Pillai/s, which lies between 0 and 1 and can be read as an effect size. Initial analyses found that residuals were often age-heteroskedastic and heavy-tailed, so we could not rely on an analytic F approximation. Instead, a null distribution was constructed using Freedman-Lane sampling (10000 draws) of reduced-model residuals with Rademacher sign flipping \cite{Winkler2014}. 
		
		A follow-up analysis was also carried out to scrutinize age-related sexual dimorphism on a per-mode basis. Similarly, each retained mode was separately modeled with the same cubic form of the age term. These models were then fitted with HC3 heteroskedasticity-consistent variance estimators \cite{Long2000}. Within each fit, we tested the sex $\times$ age term with a joint F-test and applied Benjamini-Hochberg corrections to the resulting $p$-values.

%% file: mainmatter/3_results.tex
\section{Results}

\subsection{Detection -- segmentation framework}

	Our two-stage detection-segmentation framework accurately localized and segmented the relevant bones of interest in RSNA radiographs. For the detection component, YOLOv12s performed best, with an mAP50-95 score of 0.9711 (Table 2). However, the metrics were relatively saturated, with the worst-performing network, YOLOv12n, still scoring 0.95. Whilst detection performance was generally correlated with model size, YOLOv12s still outperformed the 'm'-sized variants of YOLOv8, YOLOv11, and YOLOv12 with $\sim$11M fewer components (Figure 2).

	For our universal bone segmentation head, SegFormerB4 ranked best on our HD95 + ASSD metric, with the lowest value of 1.961 (Table 3). Decomposed, this configuration had the lowest HD95 (1.512 $\pm$ 0.005) and an ASSD of 0.449 $\pm$ 0.001. SAM2 Hiera Large, by contrast, had the best ASSD, Dice, and IoU scores of 0.358 $\pm$ 0.009, 0.989 $\pm$ 0.00012, and 0.978 $\pm$ 0.00023, respectively. However, a caveat is that SAM2 Hiera Large also required significantly more computational resources to train than SegFormerB4: 354 PFLOPs versus 45 PFLOPs. Regardless, we chose SegFormerB4 as the segmentation-head architecture based on the ranking score.  

\subsection{Statistical shape models}

	Our constructed SSMs encapsulated the overall shape space of age- and sex-varying anatomy into distinct shape modes and compact shape parameters. These shape modes captured the principal modes of variation across different hand bones (Figure 3). For all hand bones, the first 7 shape modes captured 90\% of total shape variation, except for the proximal phalanx, which required 8 modes.

	For the distal phalanx (Figure 3), the first mode is characterized by a variation in the aspect ratio of the bone, that is, an inverse correlation in the longitudinal and lateral lengths. The second mode corresponds to a shear-like deformation, characterized by correlated variations along the bone's diagonal edges. The third mode accounts for variation in the delineation between the tuft and shaft via a difference in the degree of tapering separating these regions. 
	
	For the medial phalanx (Figure 3), the first mode encompasses a delineation between two distinct morphologies characterized by correlated variation in diaphysis thickness and length, head filleting, and base convexity. The second mode reflects a shear-like deformation with correlated variations along diagonal edges. The third mode is characterized by an elongation of the head axially with a corresponding contraction of the base. 
	
	For the proximal phalanx (Figure 3), the first mode encompasses a variation in aspect ratio characterized by shaft length and thickness. This is compounded by variations in head filleting and decreased base concavity, culminating in a distinct morphological form. The second mode corresponds to head elongation with a corresponding base contraction. The third shape mode accounts for a shear-like variation previously observed in other phalanges, namely, correlated variations in opposing corners with inversely correlated variations in the remaining corners. 
	
	For the metacarpals (Figure 3), the first shape mode accounts for variations that seemingly delineate pediatric and NSM morphologies. That is, increased overall size and a rounding of the distal head and proximal base. The second shape mode accounts for the overall thickness of the bone from the head to the base. The third shape mode corresponds to a shear-like deformation described previously; however, it appears more pronounced in the head than the base.

\subsection{Sexual dimorphism} 

	Hotelling's $T^2$ metrics show that all bones of the third ray exhibit statistically significant sexual dimorphism (Table 4). Specifically, $T^2$ values range from 66.656 to 142.01, and $p < 0.001$ across all four bones. The Mahalanobis $\mathcal{D}$ values range from 0.4279 to 0.6246, indicating clear differences between male and female shape parameters. 

	We then visualized the mean male and female shapes by obtaining stratified mean shape parameters from the retained analysis modes for each bone. We then utilized these parameters to morph the mean shape of each bone following Equation \ref{eq:ssm} (Figure 4A). Overlaying these mean shapes then allowed us to intuitively visualize and quantify dimorphisms.
	
	For the distal phalanx, sexual dimorphism is most evident at the metaphysis, with minor differences in the head and base sizes. Specifically, the female mean shape has a thinner head with less pronounced tapering. For the medial phalanx, the main difference is width, with the male mean shape thicker, with a clear difference proximally. 
	
	For the proximal phalanx, sexual dimorphism is evident in base thickness, with males clearly thicker than females; the difference is also visible in diaphyseal thickness. The metacarpal, on the other hand, exhibited a relatively smaller magnitude of sexual dimorphism. The main discernible differences are in the base width and metaphysis shape. Furthermore, shaft width is similarly dimorphic. 

	On a per-mode basis, we further scrutinized the prevalence of sexual dimorphism. Welch's $t$-tests allowed us to identify, for each bone, which modes were sexually dimorphic (Table 5). We also visualized the sex-stratified shape parameters as violin plots for the first 5 shape modes (Figure 4B). Although similar, differences in stratified shape parameter distributions are discernible, and statistical testing confirms their significance. Specifically, the first shape mode was consistently significantly dimorphic across all bones, with Cohen's $d$ ranging from 0.1333 to 0.4806. That is, variations broadly relating to size and aspect ratio are significant sources of sexual dimorphism. 

\subsection{Sexual dimorphism over time}

	Results from the interaction block of our multivariate general linear model demonstrated that sexes do not follow parallel age trajectories (Table 6). $p$-values obtained via Freedman-Lane resampling highlighted significant differences across all bones. Moreover, effect sizes based on normalized Pillai's trace (Pillai/s) ranged from 0.032 -- 0.058 for the distal and proximal phalanges, respectively, which are relatively modest.
	
	Our subsequent granular joint F-tests on the interaction block for each mode highlighted specific modes where age trajectories were sexually dimorphic (Table 7, Figure 5). The distal phalanx, for example, exhibited significantly different trajectories only on PC1, PC4, PC5, and PC6. PC1, in particular, shows differing initial values that intersect and traverse from $\sim$40 to $\sim$180 months along parallel but vertically displaced trajectories before intersecting again at $\sim$180 months and diverging. For the medial phalanx, all shape modes except PC7 differed significantly. The trajectories for PC1 begin and end similarly; however, the female trajectories tend towards negative parameters in comparison to the male, that is, a tendency toward negative parameters compared with the male trajectories, \textit{i.e.,} a tendency toward thinner morphology. For PC2, the significance was less pronounced, with trajectories overlapping across most of the age range until the final months, when they diverge sharply. For PC3, the trajectories start at different points before intersecting and running together from about $\sim$70–120 months, then diverging toward later ages. For PC4, the trajectories are distinct and begin and end at diverging points, with one intersection at $\sim$140 months.
	
	The proximal phalanx showed significant differences in trajectories across all shape modes, though to varying extents. PC1, for instance, had similar start and end points; however, it showed distinct intervening trajectories between $\sim$50--200 months. PC2, by contrast, has nearly coincident trajectories until the final months, when they diverge slightly. For PC3, the trajectories originate at separate baseline values and are interwoven as they traverse the age range, resulting in distinct trajectories. For PC4, the trajectories are well coincident until $\sim$140 months, after which they diverge significantly. Finally, for the metacarpals, all shape modes differed significantly except PC2 and PC4. PC1, again, starts with distinct initial values but intersects and remains coincident from $\sim$100 months onward. The trajectories of PC3, however, began with differing origins but intersected at $\sim$60 months and remained coincident until ~160 months, thereafter diverging. 

	With established trajectories of each shape mode for each bone over time, we can estimate the mean shapes for each sex at distinct time points to visualize hallmarks of growth and development (Figure 6). We can also overlay these mean shapes to intuitively visualize and compare sexual dimorphisms over time.

	In the distal phalanx, for instance, we can observe an overall lengthening of the bone and a change in proportions. Namely, the shaft itself lengthens, and the metaphysis becomes less pronounced. The head also shrinks relative to the base, while the latter develops further to form an interfacing surface. Comparing the male and female shapes, we can observe that the latter are generally longer and thinner. Further differences are localized around the metaphysis, with the female shape exhibiting less pronounced tapering. Interestingly, at the earliest and latest time points, the magnitude of differences is minimal. 
	
	For the medial phalanx, development is characterized by overall thinning and lengthening. The head and base also develop further with a sharpening of the lateral edges to form distinct articulating surfaces. Male and female dimorphisms were relatively minor. Male bones were discernibly thicker from 84 months onward, while female bones exhibited sharper lateral edges both proximally and distally from 120 months onward. 
	
	The proximal phalanx has a rounded head and flat base at an early age. This changes with age, as the head expands laterally with a marked increase in sharpening and the formation of a concave articulating surface. The base develops similarly, with a similar concavity forming a an articulation interface. In terms of the trajectory of sexual dimorphism, the main difference is apparent in overall length and width, with male bones being thicker and shorter. Interestingly, at 204 months, we observe the greatest magnitude of dimorphism, with females showing a greater degree of tapering in the metaphysis.
	
	Finally, for the metacarpal, developmental shape changes can be characterized by general shaft elongation and a more pronounced metaphysis. Furthermore, an overall increase in size, especially in the head and base, is distinguishable. Minimal sexual dimorphisms are readily apparent, with the most dimorphic time point being the earliest (\textit{i.e.,} 36 months). In detail, dimorphisms are marked by differences in base and head elongation, as well as differences in the degree of tapering separating the base and shaft.

%% file: mainmatter/4_discussion.tex
\section{Discussion}

	
	In our detection--segmentation framework, the pipeline localized and segmented all bones of interest in our age-diverse dataset of hand radiographs. In comparison to our previous study utilizing semantic segmentation architectures, this two-stage approach resulted in improved segmentation outcomes regardless of subject age \cite{melol}. For instance, IoU and Dice scores were generally improved without the need for age-stratified training data. Specifically, approaches in our previous study achieved roughly 0.899 $\pm$ 0.035 Mean IoU for semantic segmentation, whilst this approach achieved up to 0.975 $\pm$ 0.001 IoU. Homayounfar \textit{et al.} also attempted a similar semantic segmentation task and achieved up to 0.906 (CI $\pm$ 0.0075) IoU for a near-skeletally mature subset of the same RSNA dataset we utilized. Both studies are not directly comparable to ours because of the task difference (\textit{i.e.,} semantic vs. binary segmentation) and data differences; nevertheless, the comparison suggests that our two-stage approach may be a better suited to multi-object segmentation than semantic segmentation architectures. These performance improvements arose from relatively lightweight fine-tuning while leveraging pre-trained backbones, both of which may have contributed to the results. A notable observation was that, in general, larger models (\textit{i.e.,} those with more trainable parameters) did not necessarily outperform smaller ones in either detection or segmentation. Test performance is not necessarily monotonic in model size \cite{Nakkiran2021}, and given the relatively simple structure of the task and the modest size of the dataset, the additional capacity of the larger models may have offered little benefit while increasing their propensity to overfit. Thus, these results justify the utility of specialized, task-specific lightweight models. Whilst off-the-shelf visual foundation models could serve as a basis for assisted semi-automatic dataset labeling or human-in-the-loop style frameworks \cite{Homayounfar2024}, lightweight models could potentially require fewer computational resources to train and result in better performance. Regardless, our two-stage framework demonstrates exceptional segmentation performance across subjects' ages, presenting a potential direction for further development of age-agnostic medical image analysis. 

	
	Our fitted SSMs were able to encapsulate the envelope of shape variation for 1456 unique subjects across sex and age for 19 bones into compact shape modes and parameters. In examining the principal modes of variation, we observe distinct yet remarkable similarities across the bones of the third ray. In particular, the first shape modes generally corresponded to salient morphological changes most likely associated with growth and its entailed development. For instance, aspect ratio changes and the formation of distinct articular interfaces occurred both proximally and distally. As de Vries \textit{et al.} note, age-related changes generally dominate the first modes of variation in SSMs because they entail distinct shape variations \cite{deVries2025}. Thus, the use of age- and sex-stratified SSMs could be warranted when downstream applications are targeted for specific demographic subgroups.
		
	
	Our analyses extracted overall sexual dimorphism considering an age range encompassing significant developmental changes. Existing studies have only highlighted dimorphisms via post-hoc statistical analyses of osteometric measurements from manually labeled radiographs or cadaveric samples \cite{DeSilva2014, Scheuer1993, Waidhofer2015}. Instead, our approach discerns sexual dimorphism directly from statistical shape model subspaces. That is, by analyzing shape modes and their constituent parameters that describe the overarching shape space. This provides a more quantitative mechanism for analysis and appends dimorphism assessment as an auxiliary component to general SSM analyses of a population. Furthermore, using SSMs allows dimorphism findings to consider overall shape by accounting for covariance across the entire bone structure rather than individual measurements, which are myopic and limited in their ability to capture covariance. Moscoloni \textit{et al.} applied similar shape-parameter analysis and found significant sexual dimorphism in human heart shape \cite{Moscoloni2025}. 
	
	The shape modes and their sex-stratified parameters were significantly different across all bones, highlighting that all bones of the hand were sexually dimorphic (Appendix C). The dimorphisms we identified align with observations from existing morphometric studies. For the proximal phalanges, our observations that dimorphisms are most evident in terms of base width and diaphysis thickness are in line with existing literature \cite{Mahakkanukrauh2013, Karakostis2017, Karakostis2013}. Similarly, in the metacarpals, our findings that dimorphisms are most pronounced in overall thickness, especially at the base, agree with existing studies \cite{Manolis2009, Khanpetch2012}. Lastly, for the medial and distal phalanges, Rad \textit{et al.} found that proximal and distal widths were sexually dimorphic, which agrees with our observations \cite{Rad2022}.
		
	An interesting caveat in our comparisons with the literature is that these studies are generally focused on skeletally mature subjects (\textit{i.e.,} >17 years old). In contrast, the dataset we used spans the pediatric--NSM age range, a period of significant morphological change. Nevertheless, sexual dimorphism was readily discernible within this age range, with their form aligning with that found in skeletally mature subjects. Regardless, discerning sex from pediatric radiographs is a relatively non-trivial task, and, to the best of our knowledge, no established protocols exist. Yune \textit{ et al.} developed a deep learning-based classification model which outperformed clinical experts in classifying male and female radiographs in an age-diverse dataset \cite{Yune2018}. A key observation in their study is that, even with the assistance of class activation maps, human radiographers could not identify sexually distinguishing features. Thus, whilst their work demonstrates a valid approach, the lack of explainability is contentious in the context of black-box solutions for medical imaging \cite{Amann2020}. Thus, our findings provide a data-driven alternative in which pediatric sexual dimorphisms can be highlighted and quantified through unambiguous shape-parameter variations. 
	
	However, when examining the distributions of the sex-stratified shape parameters, the differences appear relatively subtle in structure and effect size (Cohen's $d = 0.25 \pm 0.104$) (Figure 4B, Table 5, Appendix C). Statistical significance notwithstanding, this could indicate that sexual dimorphisms of these bones are subtle but nevertheless remain an important consideration. These subtleties could also stem from our scaling procedure in SSM construction, which may obscure size, a major source of dimorphism \cite{Seeman2001}. Moreover, additional literature concerning sexual dimorphism of finger bones was principally focused on linear dimensions (\textit{i.e.,} length), which are likewise obscured with scaling \cite{AlFartusie2020, ElMorsi2013, KerWoon2026}

	
	Developmental shape changes in the hand bones are a well-documented phenomenon used to determine age, and are sex-delineated. The two main methodologies are the Tanner-Whitehouse (TW3) and the Greulich and Pyle (GP) methods \cite{tanner2001assessment, greulich1959radiographic, gaskin2011skeletal}. Briefly, in the latter, a clinical expert compares a target image to a set of age-delineated standards to identify the nearest match. Based on this match, an estimate of skeletal age can be obtained. The former is more complex, wherein particular regions of interest are scrutinized for specific morphological features and assigned a score. The scores can then be tallied and used to estimate skeletal age. When comparing both methods, GP is generally preferred because of its simplicity and ease of use; however, TW3 is more accurate but time-intensive \cite{Tewattanarat2026}. Comparing our observed morphological changes with these methods, we find striking similarities. In the metacarpals, both methods show similar trends in lateral head growth, leading to a rounding and an increased diameter that surpasses that of the diaphysis. For the proximal phalanx, the base similarly begins relatively flat, leading to a concave interface after the capping and fusion of the epiphysis. The medial phalanx shows similar changes overall, with the base starting flat and developing concavities after epiphyseal capping and fusion. Likewise, the distal phalanx shows similar changes in the proximal base, with an articular interface forming after epiphyseal capping and fusion. Interestingly, the relative changes in bone proportions are also visible: the head and base are similar in diameter at an early age, but in later ages the base is markedly wider. 
	
	These comparisons are, in general, qualitative due to the nature of the methods themselves. Nevertheless, they highlight both the validity of our study and the shortcomings of existing methods. Stage descriptors for TW3 and reference standards for GP describe changes and morphology very subjectively. For example, for TW3, stage E for the third metacarpal is described as “…the shape of the epiphyses has changed from being an oval or semicircle to that of a spade or fingernail...”. In GP, evaluations rely on the accuracy of the reference standard and subjective interpretation to assess similarities. Overall, both require the interpretation of a clinical expert, which could be time- and resource-intensive, whilst also lacking distinct interpretation, which can be time- and resource-intensive, while also lacking clear quantitative objectivity. Both methods also rely on non-diverse reference populations, which has motivated some researchers to recommend ethnicity-based corrections \cite{Yuh2023}. However, alternative studies across ethnic groups have found inconsistent results that depend heavily on sex and specific age groups, warranting further research \cite{Pinchi2014, KowoNyakoko2023}. Our approach, in contrast, provides a pathway to derive unambiguous data-driven descriptors and quantify population-based developmental shape changes objectively.  
	
 
	Our study was limited in two ways: data and methodology. In terms of the data, the RSNA dataset used in this study was limited to 2D radiographs, with metadata limited to subjects' sex and skeletal age. An interesting extension would be to examine discrepancies between skeletal and chronological age and their relation to discernible morphological variation. Furthermore, as data provenance is generally unknown, we cannot investigate or control for the effects of other potential confounding variables (\textit{e.g.,} ethnicity, physical activity, biomarker levels, \textit{etc.}) similar to the work of Moscoloni \textit{et al.} \cite{Moscoloni2025}. Because the data are also limited to 2D, we cannot map the full morphological variations, which may underrepresent the overall shape space. Applying the methodology developed in this study to full 3D imaging could uncover and map said shape space in greater detail. Furthermore, having image acquisition metadata would enable a direct correlation to physical units, which could yield additional clinical and anthropological insights. The lack of acquisition metadata also precluded the use of unscaled SSMs and necessitated scaling of the landmarks. As discussed, this could have potentially obscured a distinct source of variation and warrants further study. 
	
	
	Lastly, our use of PDM-based SSMs also presented a limitation in the structures that could be modeled. Our bone masks generally excluded epiphyses for all bones unless they capped the diaphysis. This was done as capping of the epiphyses has been found to be indicative of a cessation of growth and generally denotes the final length of the bone \cite{Parfitt2002}. Nevertheless, the TW3 method generally focuses on epiphyseal morphology and growth dynamics as an indicator of skeletal age. Our approach, whilst still highlighting sexual dimorphisms and sexually dimorphic developmental shape changes, could be inadvertently discounting valuable radiographic features and warrants further study. Potentially, multi-body SSMs could jointly model the diaphysis and unfused epiphyses. However, placing consistent landmarks across appearing and disappearing anatomy (\textit{i.e.,} epiphyseal fusion) is non-trivial.  Nevertheless, this suggests that alternative SSMs could be developed or used for similar spatiotemporal shape-modeling studies, and it remains an open avenue for exploration.

%% file: mainmatter/5_conclusion.tex
\section{Conclusions}

	In conclusion, our work sought to chart the overarching envelope of anatomical shape variation across distinct age groups for hand and finger bones from 2D radiographs. We used SSMs and post-hoc analyses of stratified shape parameters to quantitatively characterize sexual dimorphism and sex-based developmental shape changes. Our findings illustrated these dimorphisms across a population scale and within age-specific subgroups in an intuitive way, with the added benefit of operating within an unambiguous, quantitative framework. Overall, this work lays the foundation for further population-scale analyses of growing biological tissue. Our insights into the utility of SSM-based methods, which distill population-scale variation into a compact subspace, could be amalgamated with longitudinal data to inform patient-specific prediction models, warranting a potential direction of exploration. Further work is also necessary to explore the applicability of this method to alternative biological tissues and imaging modalities.

%% file: mainmatter/6_competing.tex
\section*{Competing Interests}

	\noindent
	The authors declare no competing interests. 

%% file: mainmatter/8_acknowledgements.tex
\section*{Acknowlegements}

	\noindent 
	The authors would like to thank the Radiological Society of North America for publishing the RSNA Pediatric Bone Age Challenge 2017 image dataset as open source, which we used in this study. This dataset is available at the following link (https://www.rsna.org/education/ai-resources-and-training/ai-image-challenge/RSNA-Pediatric-Bone-Age-Challenge-2017).

%% file: mainmatter/9b_table_captions.tex
\pagebreak

\section*{Table Captions}

	\paragraph{Table 1.} Dataset demographics for the full RSNA dataset and the subset extracted for this study.
	
	\begin{table}[h!]
		\centering
		\resizebox{\textwidth}{!}{%
		\begin{tabular}{l|l|lllll|l}
		\hline
 		& \multirow{2}{*}{Sex} & \multicolumn{5}{c|}{Age (months)} & \multirow{2}{*}{Total} \\ \cline{3-7}
 		&  & 0 - 43.2 & 43.2 - 86.4 & 86.4 - 129.6 & 129.6 - 172.8 & 172.8 - 230 &  \\ \hline
		\multicolumn{1}{c|}{\multirow{2}{*}{Full RSNA}} & Male & 233 & 894 & 1164 & 3690 & 852 & 6833 \\
		\multicolumn{1}{c|}{} & Female & 249 & 970 & 1986 & 2223 & 350 & 5778 \\
		\multirow{2}{*}{Analysis subset} & Male & 118 & 143 & 151 & 156 & 156 & 724 \\
 		& Female & 119 & 145 & 152 & 157 & 159 & 732 \\ \hline
		\end{tabular}%
		}
		\label{tb:datademographics}
	\end{table}
	\FloatBarrier
	
	\paragraph{Table 2.} Results from detector model sweep. A 200-epoch cap was set with patience 15, all runs early-stopped so \texttt{Epochs} indicates the number of epochs actually run. Highest mAP50-90 is indicated in bold.

	\begin{table}[h!]
		\centering
		\resizebox{\ifdim\width>\textwidth \textwidth \else \width \fi}{!}{%
		\begin{tabular}{l|l|lll}
		\hline
		Model & Size & Params (M) [\#] & Epochs [\#] & mAP50-95 ($\uparrow$) [-] \\
		\hline
		\multirow{3}{*}{YOLOv8} & n & 3.01 & 82 & 0.9526 \\
		 & s & 11.14 & 79 & 0.9622 \\
		 & m & 25.87 & 122 & 0.9705 \\ \cline{1-5}
		\multirow{3}{*}{YOLO11} & n & 2.59 & 81 & 0.9521 \\
		 & s & 9.44 & 94 & 0.9663 \\
		 & m & 20.07 & 76 & 0.9676 \\ \cline{1-5}
		\multirow{3}{*}{YOLO12} & n & 2.57 & 68 & 0.9500 \\
		 & s & 9.26 & 124 & \textbf{0.9711} \\
		 & m & 20.15 & 69 & 0.9674 \\
		\hline
		\end{tabular}
		}
		\label{tab:detector_sweep}	
	\end{table}
	\FloatBarrier

	\paragraph{Table 3.} Results from the universal bone segmentation head sweep across model architectures and encoder backbones. We trained each configuration in triplicate with randomized seeds for robustness, and report results as mean $\pm$ standard deviation. For each metric, the best values are highlighted in bold. 

\begin{table}[h!]
    \centering
    \resizebox{\ifdim\width>\textwidth \textwidth \else \width \fi}{!}{%
    \begin{tabular}{ll|ll|lllll}
    \hline
        Head & Encoder & Params [M] & Train [PFLOPs] & HD95 + ASSD ($\downarrow$) [px] & HD95 ($\downarrow$) [px] & ASSD ($\downarrow$) [px] & Dice ($\uparrow$) [-] & IoU ($\uparrow$) [-] \\
    \hline
        \multirow{6}{*}{SegFormer} & mit\_b0 & 3.7 & 6 & 2.027 & $1.564 \pm 0.010$ & $0.463 \pm 0.001$ & $0.98683 \pm 0.00003$ & $0.97408 \pm 0.00005$ \\
         & mit\_b1 & 13.7 & 12 & 1.988 & $1.531 \pm 0.009$ & $0.457 \pm 0.001$ & $0.98698 \pm 0.00003$ & $0.97438 \pm 0.00006$ \\
         & mit\_b2 & 24.7 & 15 & 1.978 & $1.526 \pm 0.015$ & $0.452 \pm 0.003$ & $0.98713 \pm 0.00007$ & $0.97467 \pm 0.00015$ \\
         & mit\_b3 & 44.6 & 29 & 1.977 & $1.524 \pm 0.013$ & $0.453 \pm 0.003$ & $0.98711 \pm 0.00005$ & $0.97462 \pm 0.00010$ \\
         & mit\_b4 & 61.4 & 42 & \textbf{1.962} & $\mathbf{1.512 \pm 0.005}$ & $0.449 \pm 0.001$ & $0.98718 \pm 0.00003$ & $0.97476 \pm 0.00006$ \\
         & mit\_b5 & 82.0 & 45 & 2.003 & $1.548 \pm 0.025$ & $0.455 \pm 0.003$ & $0.98705 \pm 0.00006$ & $0.97451 \pm 0.00012$ \\ \cline{1-9}
        \multirow{4}{*}{SAM2} & hiera\_tiny & 39.0 (4.22) & 34 & 2.044 & $1.669 \pm 0.108$ & $0.374 \pm 0.010$ & $0.98820 \pm 0.00007$ & $0.97679 \pm 0.00014$ \\
         & hiera\_small & 46.1 (4.22) & 48 & 2.077 & $1.693 \pm 0.056$ & $0.383 \pm 0.008$ & $0.98795 \pm 0.00017$ & $0.97631 \pm 0.00032$ \\
         & hiera\_base\_plus & 80.9 (4.22) & 127 & 2.185 & $1.806 \pm 0.291$ & $0.379 \pm 0.038$ & $0.98855 \pm 0.00033$ & $0.97748 \pm 0.00064$ \\
         & hiera\_large & 224.4 (4.22) & 354 & 1.991 & $1.633 \pm 0.056$ & $\mathbf{0.358 \pm 0.009}$ & $\mathbf{0.98885 \pm 0.00012}$ & $\mathbf{0.97804 \pm 0.00023}$ \\ \cline{1-9}
        \multirow{5}{*}{DeepLabV3+} & resnet18 & 12.3 & 14 & 2.071 & $1.590 \pm 0.017$ & $0.481 \pm 0.002$ & $0.98632 \pm 0.00002$ & $0.97310 \pm 0.00005$ \\
         & resnet34 & 22.4 & 20 & 2.074 & $1.597 \pm 0.021$ & $0.477 \pm 0.004$ & $0.98643 \pm 0.00007$ & $0.97332 \pm 0.00013$ \\
         & resnet50 & 26.7 & 35 & 2.038 & $1.569 \pm 0.026$ & $0.469 \pm 0.004$ & $0.98663 \pm 0.00008$ & $0.97370 \pm 0.00015$ \\
         & resnet101 & 45.7 & 53 & 2.079 & $1.605 \pm 0.060$ & $0.474 \pm 0.009$ & $0.98657 \pm 0.00019$ & $0.97359 \pm 0.00036$ \\
         & resnet152 & 61.3 & 74 & 1.999 & $1.537 \pm 0.022$ & $0.462 \pm 0.002$ & $0.98680 \pm 0.00006$ & $0.97402 \pm 0.00012$ \\ \cline{1-9}
        \multirow{1}{*}{nnU-Net} & -- & 20.6 & 565 & 2.115 & $1.651 \pm 0.015$ & $0.464 \pm 0.002$ & $0.98701 \pm 0.00006$ & $0.97448 \pm 0.00010$ \\ \cline{1-9}
        \multirow{1}{*}{MedSAM2} & hiera\_t512 & 39.0 (4.22) & 8 & 2.333 & $1.903 \pm 0.036$ & $0.430 \pm 0.007$ & $0.98657 \pm 0.00021$ & $0.97365 \pm 0.00042$ \\
    \hline
    \end{tabular}
    }
    \label{tab:segmentation_head_sweep}
\end{table}
	\FloatBarrier

	\paragraph{Table 4.} Hotelling's $T^2$ over sex-stratified shape parameters encompassing $c$ shape modes retained for analysis. $F$ and $p$-values denote statistical significance whilst $T^2$ and $\mathcal{D}$ quantifies the distance between males and females.
	
		\begin{table}[h!]
		\centering
		\resizebox{\ifdim\width>\textwidth \textwidth \else \width \fi}{!}{%
		\begin{tabular}{l|lllllll}
		\hline
		Bone & $T^2$ & $F$ & df & $\mathcal{D}$ & $p$-value \\
		\hline
		Distal Phalanx 3 & 66.656 & 9.483 & 7, 1448 & 0.4279 & $p \leq 0.001$ \\
		Medial Phalanx 2 & 137.89 & 19.62 & 7, 1448 & 0.6155 & $p \leq 0.001$ \\
		Proximal Phalanx 3 & 142.01 & 17.67 & 8, 1447 & 0.6246 & $p \leq 0.001$ \\
		Metacarpal 3 & 71.562 & 10.18 & 7, 1448 & 0.4434 & $p \leq 0.001$ \\
		\hline
		\end{tabular}
		}
		\label{tab:dimorphism_omnibus}
	\end{table}
	\FloatBarrier

	\paragraph{Table 5.} Per-mode sexual dimorphism tested via Welch's $t$-tests. $t$-statistic and Cohen's $d$ values quantify the differences in groups whilst $p$-values denote significance. $p$-values are corrected with the Benjamini-Hochberg (BH) method and denote significance if $p \leq 0.05$. $p \geq 0.05$ are marked as not significant (\texttt{ns}).
	
		\begin{table}[h!]
		\centering
		\resizebox{\ifdim\width>\textwidth \textwidth \else \width \fi}{!}{%
		\begin{tabular}{ll|lll}
		\hline
		Bone & mode & $t$ & Cohen's $d$ & $p$-value (BH) \\
		\hline
		\multirow{7}{*}{Distal phalange 3} & PC1 & 6.604 & 0.3462 & $p \leq 0.001$ \\
 		& PC2 & 4.63 & 0.243 & $p \leq 0.001$ \\
 		& PC3 & $-$0.7269 & $-$0.03811 & ns \\
 		& PC4 & 1.123 & 0.05891 & ns \\
 		& PC5 & 1.578 & 0.08273 & ns \\
 		& PC6 & $-$0.6306 & $-$0.03305 & ns \\
 		& PC7 & 2.31 & 0.1211 & $p \leq 0.05$ \\ \cline{1-5}
		\multirow{7}{*}{Medial phalange 2} & PC1 & 7.377 & 0.3867 & $p \leq 0.001$ \\
 		& PC2 & $-$1.213 & $-$0.06356 & ns \\
 		& PC3 & $-$7.828 & $-$0.4106 & $p \leq 0.001$ \\
 		& PC4 & $-$2.593 & $-$0.1359 & $p \leq 0.05$ \\
 		& PC5 & $-$0.169 & $-$0.008859 & ns \\
 		& PC6 & $-$4.376 & $-$0.2295 & $p \leq 0.001$ \\
 		& PC7 & $-$2.204 & $-$0.1156 & $p \leq 0.05$ \\ \cline{1-5}
		\multirow{8}{*}{Proximal phalange 3} & PC1 & 9.17 & 0.4806 & $p \leq 0.001$ \\
 		& PC2 & $-$2.028 & $-$0.1063 & ns \\
 		& PC3 & $-$0.03245 & $-$0.001701 & ns \\
 		& PC4 & 4.355 & 0.2283 & $p \leq 0.001$ \\
 		& PC5 & $-$0.5092 & $-$0.02668 & ns \\
 		& PC6 & $-$0.8242 & $-$0.04321 & ns \\
 		& PC7 & $-$2.548 & $-$0.1336 & $p \leq 0.05$ \\
 		& PC8 & $-$4.371 & $-$0.2293 & $p \leq 0.001$ \\ \cline{1-5}
		\multirow{7}{*}{Metacarpal 3} & PC1 & $-$2.538 & $-$0.1333 & $p \leq 0.05$ \\
		 & PC2 & $-$7.59 & $-$0.3979 & $p \leq 0.001$ \\
		 & PC3 & $-$1.777 & $-$0.09324 & ns \\
		 & PC4 & 1.903 & 0.09989 & ns \\
		 & PC5 & $-$2.676 & $-$0.1403 & $p \leq 0.05$ \\
		 & PC6 & 0.7702 & 0.04041 & ns \\
		 & PC7 & $-$1.696 & $-$0.08896 & ns \\
		\hline
		\end{tabular}
		}
		\label{tab:dimorphism_per_mode}
	\end{table}
	\FloatBarrier

	\paragraph{Table 6.} Results from Pillai's trace on sex $\times$ age interaction. Raw Pillai's trace magnitudes are normalized with respect to $s$ degrees of freedom, giving Pillai$/s$. We report raw $p$-values based on analytic $F$, but because residuals are heteroskedastic, we used permutation tests via Freedman-Lane sampling with Rademacher sign flipping to obtain a more robust indicator of significance.
	
	\begin{table}[h!]
		\centering
		\resizebox{\ifdim\width>\textwidth \textwidth \else \width \fi}{!}{%
		\begin{tabular}{l|lllllll}
		\hline
		Bone & Pillai & $s$ & Pillai$/s$ & $F$ & df & $p$ ($F$) & $p$ (Freedman--Lane) \\
		\hline
		Distal Phalanx 3 & 0.09712 & 3 & 0.03237 & 6.902 & 21, 4332 & $p \leq 0.001$ & $p \leq 0.001$ \\
		Medial Phalanx 2 & 0.1231 & 3 & 0.04104 & 8.828 & 21, 4332 & $p \leq 0.001$ & $p \leq 0.001$ \\
		Proximal Phalanx 3 & 0.1726 & 3 & 0.05752 & 11.01 & 24, 4329 & $p \leq 0.001$ & $p \leq 0.001$ \\
		Metacarpal 3 & 0.1498 & 3 & 0.04995 & 10.84 & 21, 4332 & $p \leq 0.001$ & $p \leq 0.001$ \\
		\hline
		\end{tabular}
		}
		\label{tab:interaction_omnibus}
	\end{table}	
	\FloatBarrier
	
	\paragraph{Table 7.} Per-mode sex $\times$ age analysis. Results from the joint F-test highlighted, for each mode, whether shape trajectories were sexually dimorphic. $p$-values are corrected with the Benjamini-Hochberg (BH) method and denote significance if $p \leq 0.05$. $p \geq 0.05$ are marked as not significant (\texttt{ns}).
	
		\begin{table}[h!]
		\centering
		\resizebox{\ifdim\width>\textwidth \textwidth \else \width \fi}{!}{%
		\begin{tabular}{l|l|lll}
		\hline
		Bone & mode & $F$ & df & $p$-value (BH) \\
		\hline
		\multirow{7}{*}{Distal Phalanx 3} & PC1 & 22.93 & 3, 1448 & $p \leq 0.001$ \\
		 & PC2 & 0.7691 & 3, 1448 & ns \\
		 & PC3 & 1.357 & 3, 1448 & ns \\
		 & PC4 & 7.259 & 3, 1448 & $p \leq 0.001$ \\
		 & PC5 & 4.457 & 3, 1448 & $p \leq 0.01$ \\
		 & PC6 & 12.69 & 3, 1448 & $p \leq 0.001$ \\
		 & PC7 & 1.336 & 3, 1448 & ns \\ \cline{1-5}
		\multirow{7}{*}{Medial Phalanx 2} & PC1 & 5.631 & 3, 1448 & $p \leq 0.01$ \\
		 & PC2 & 5.271 & 3, 1448 & $p \leq 0.01$ \\
		 & PC3 & 15.44 & 3, 1448 & $p \leq 0.001$ \\
		 & PC4 & 10.87 & 3, 1448 & $p \leq 0.001$ \\
		 & PC5 & 10.74 & 3, 1448 & $p \leq 0.001$ \\
		 & PC6 & 10.69 & 3, 1448 & $p \leq 0.001$ \\
		 & PC7 & 0.2192 & 3, 1448 & ns \\ \cline{1-5}
		\multirow{8}{*}{Proximal Phalanx 3} & PC1 & 5.534 & 3, 1448 & $p \leq 0.01$ \\
		 & PC2 & 3.541 & 3, 1448 & $p \leq 0.05$ \\
		 & PC3 & 6.025 & 3, 1448 & $p \leq 0.001$ \\
		 & PC4 & 17.99 & 3, 1448 & $p \leq 0.001$ \\
		 & PC5 & 3.907 & 3, 1448 & $p \leq 0.05$ \\
		 & PC6 & 12.07 & 3, 1448 & $p \leq 0.001$ \\
		 & PC7 & 3.291 & 3, 1448 & $p \leq 0.05$ \\
		 & PC8 & 21.82 & 3, 1448 & $p \leq 0.001$ \\ \cline{1-5}
		\multirow{7}{*}{Metacarpal 3} & PC1 & 13.38 & 3, 1448 & $p \leq 0.001$ \\
		 & PC2 & 0.5082 & 3, 1448 & ns \\
		 & PC3 & 2.993 & 3, 1448 & $p \leq 0.05$ \\
		 & PC4 & 1.393 & 3, 1448 & ns \\
		 & PC5 & 12.52 & 3, 1448 & $p \leq 0.001$ \\
		 & PC6 & 17.02 & 3, 1448 & $p \leq 0.001$ \\
		 & PC7 & 14.86 & 3, 1448 & $p \leq 0.001$ \\
		\hline
		\end{tabular}
		}
		\label{tab:interaction_per_mode}
	\end{table}
	\FloatBarrier

	\vfill

%% file: mainmatter/9a_figure_captions.tex
\pagebreak

\section*{Figure Captions}

	\paragraph{Figure 1.} Overview of the analysis framework employed within this study to obtain bone image masks and build statistical shape models (SSMs). A) A subset of data is randomly extracted from the RSNA Bone Age challenge dataset \cite{Halabi2019}, whilst ensuring relatively equal distribution across sexes and ages. B) The raw data is fed into a trained detector to obtain bounding boxes for each bone. Each bounding box is then fed into a trained segmentation head to obtain binary image masks of the bone of interest within each box. These masks are then collected to obtain image masks for each finger bone of each subject. C) The masks are then fed into an SSM pipeline. There, the data are first aligned, landmarks are propagated from a template bone to each bone in the dataset, and principal component analysis (PCA) is performed. 
 
	\paragraph{Figure 2.} Evaluation of different detection models (YOLOv8, YOLOv11, YOLOv12) and their sizes (n, s, m, in ascending order) in terms of performance against the number of trainable parameters. Performance is evaluated by mean average precision across 10 different intersection-over-union thresholds, from 50\% to 95\% (mAP50-95). Note for this metric, higher is better. 

	\paragraph{Figure 3.} Statistical shape models (SSMs) of the third-ray bones, shown as their mean shape and the first 3 shape modes. Presented alongside each SSM are plots showing the cumulative variance captured by each shape mode, alongside an analysis threshold (set at 90\%). 
	
	\paragraph{Figure 4.} A) Sex-stratified mean shapes for each bone obtained by morphing their respective mean shapes by mean sex-stratified shape parameters. Differences are calculated for each landmark and plotted as a heatmap highlighting dimorphisms on the contours themselves. B) Visualization of the sex-stratified shape parameters of the first 5 shape modes for each bone. Significance is determined via Welch's $t$-tests and denoted with asterisks. $^{*}-p < 0.05$, $^{**}-p<0.01$, $^{***}-p<0.001$.

	\paragraph{Figure 5.} Shape parameters for the first five shape modes of each subject of the third ray are plotted as a function of skeletal age. Overlaid are sex-specific trajectories fitted based on a multivariate general linear model. Significance is determined via joint F-tests with Benjamini-Hochberg corrections and denoted with asterisks. $^{*}-p < 0.05$, $^{**}-p<0.01$, $^{***}-p<0.001$. 
	
	\paragraph{Figure 6.} Visualization of estimated mean shapes for bones of the third ray at different skeletal ages. Estimates were obtained by predicting shape parameters from fitted multivariate general linear models at different time points for both sexes. Mean shapes were then overlaid and differences highlighted. 

	\vfill

%% file: mainmatter/10_figures.tex
\pagebreak 

\section*{Figure 1}
	\begin{figure}[h!]
		\centering
		\includegraphics{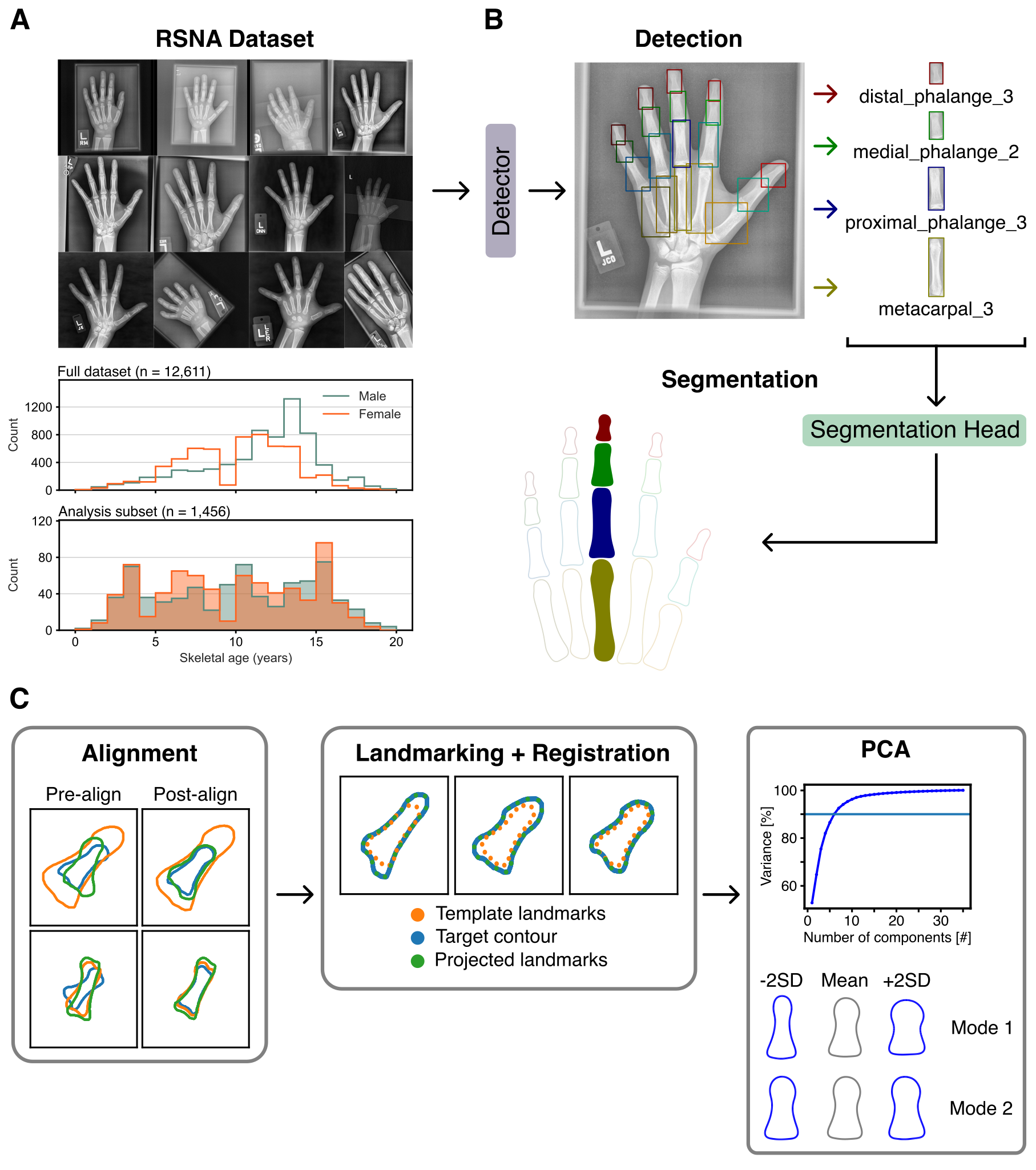}
		\label{fig:1}
	\end{figure}
 \vfill

\pagebreak 

\section*{Figure 2}

	\begin{figure}[h!]
		\centering
		\includegraphics{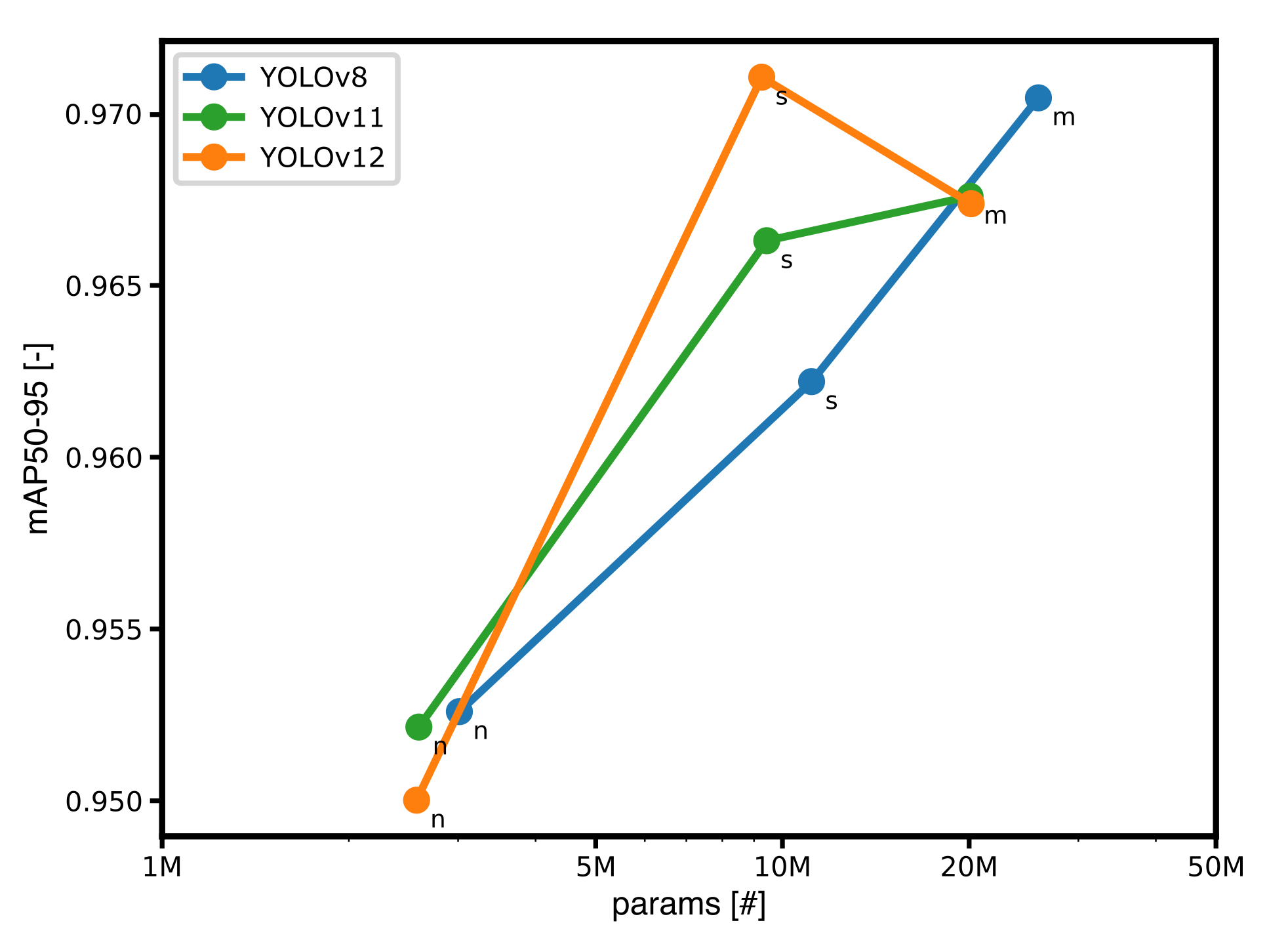}
	\end{figure}
 \vfill
\pagebreak 		

%

\section*{Figure 3}
	\begin{figure}[h!]
		\centering
		\includegraphics{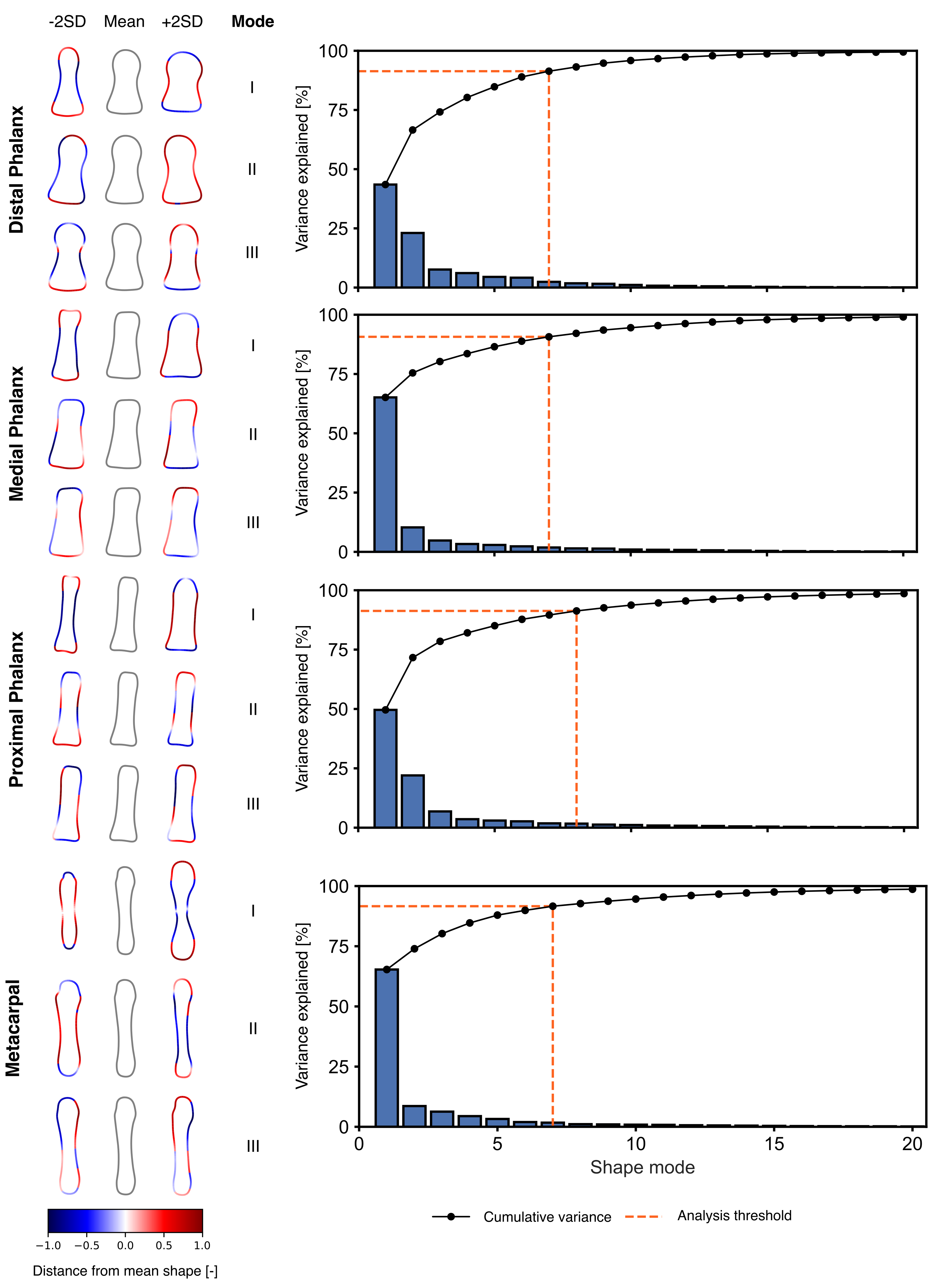}
	\end{figure}
\vfill
\pagebreak 

\section*{Figure 4}

	\begin{figure}[h!]
		\centering
		\includegraphics{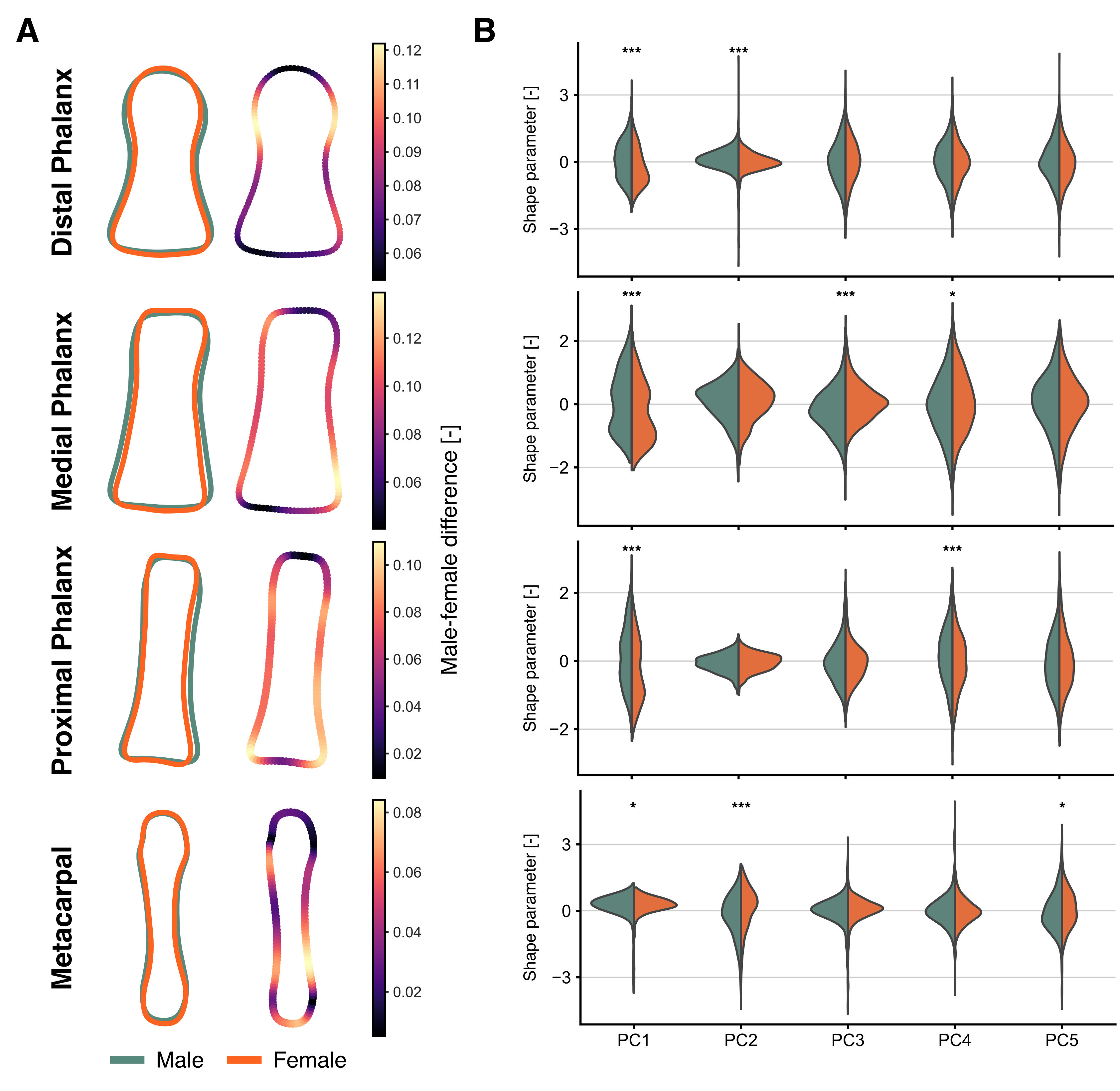}
	\end{figure}
\vfill
\pagebreak

\section*{Figure 5}
	\begin{figure}[h!]
		\centering
		\includegraphics{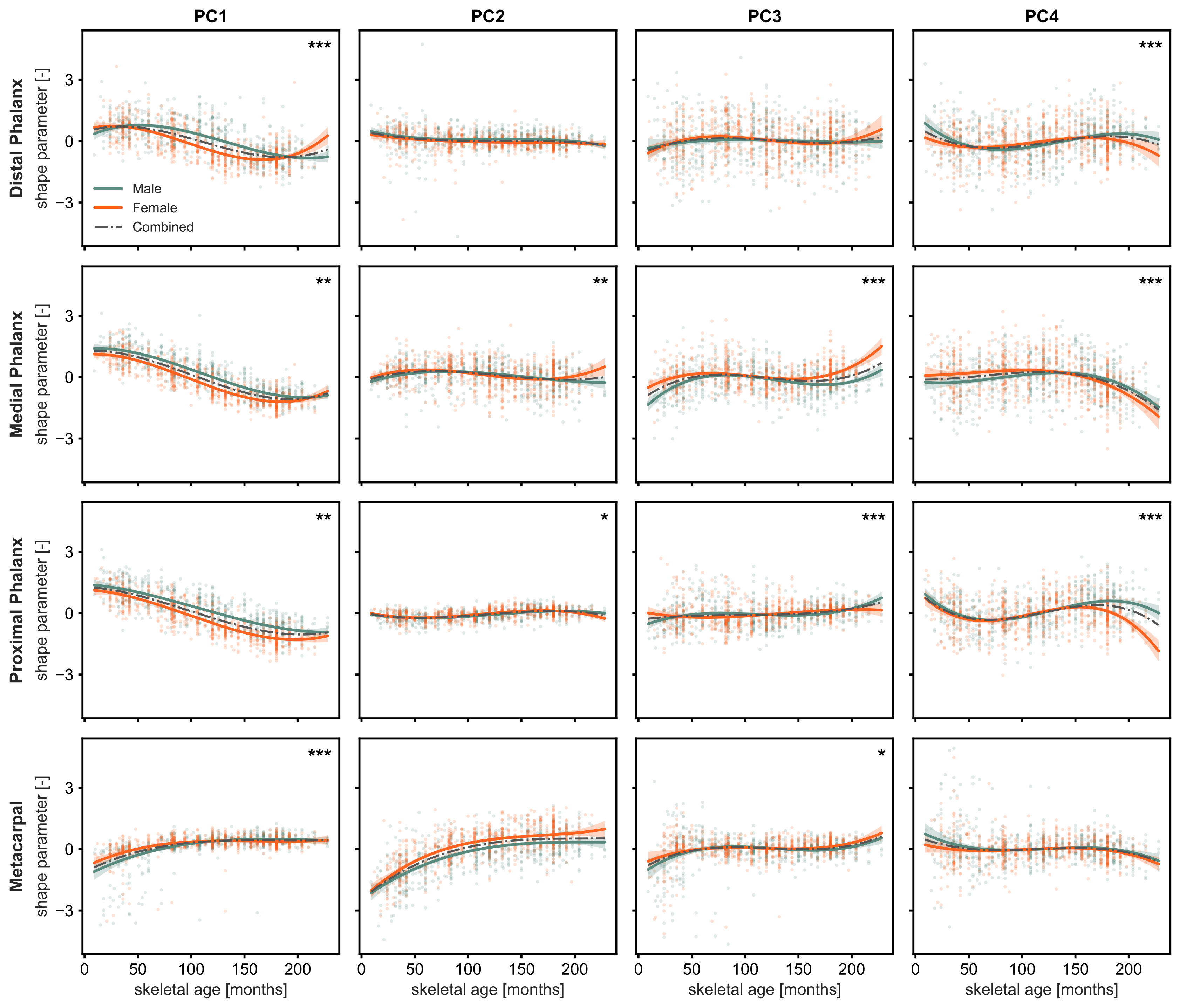}
		\label{fig:1}
	\end{figure}
 \vfill

\pagebreak 

\section*{Figure 6}

	\begin{figure}[h!]
		\centering
		\includegraphics{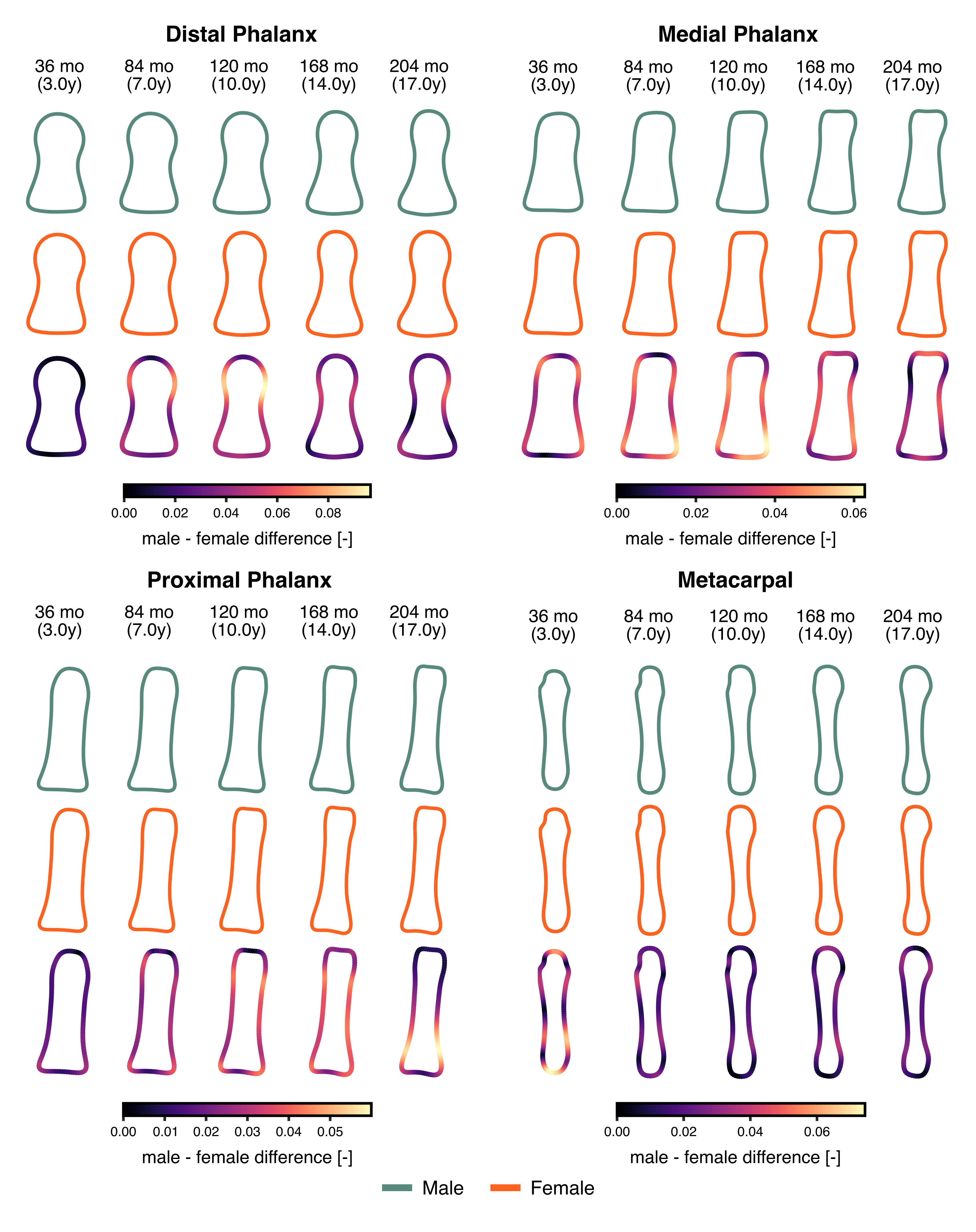}
	\end{figure}
 \vfill
\pagebreak

%% file: mainmatter/7_supplementary.tex
\newpage

\section*{Appendix}

\appendix
\renewcommand{\thetable}{\Alph{section}\arabic{table}}
\counterwithin*{table}{section}  
	
\section{Form-fitting for age term}	
	
	Because we found a significant relationship between skeletal age and hand morphology, as indicated by shape parameter variation, we characterized the form of this relationship. We fit alternative functional forms to the shape parameters for each shape mode of each bone in the third ray (Figure B1, Table B1). We then evaluated how well each model fit the parameters by calculating the Bayesian information criterion (BIC) \cite{Neath2011}. Our results showed that a cubic form fit best across all shape modes and bones. Thus, we selected this form as the representative age-term form used later in the study.  	
	
	\begin{table}[h!]
\centering
\resizebox{\ifdim\width>\textwidth \textwidth \else \width \fi}{!}{%
\begin{tabular}{l|llll}
\hline
mode & Metacarpal 3 & Proximal phalange 3 & Medial phalange 2 & Distal phalange 3 \\
\hline
PC1 & cub +1.5\textasciitilde{} & cub +14.9 & cub +52.2 & cub +34.9 \\
PC2 & quad +1.1\textasciitilde{} & cub +52.5 & cub +6.8 & lin +1.6\textasciitilde{} \\
PC3 & cub +26.5 & lin +1.1\textasciitilde{} & cub +64.0 & lin +0.1\textasciitilde{} \\
PC4 & cub +4.9 & cub +79.2 & cub +2.9 & cub +20.9 \\
PC5 & cub +53.1 & quad +4.8 & lin +7.2 & quad +7.1 \\
PC6 & cub +107.4 & cub +17.1 & cub +45.2 & cub +15.4 \\
PC7 & cub +0.6\textasciitilde{} & cub +4.4 & quad +6.8 & lin +5.0 \\
PC8 & -- & cub +5.3 & -- & -- \\
\hline
\end{tabular}
}
\caption{Age term elected per shape mode for bones of the third ray. Each cell names the lowest-BIC candidate for that bone and mode, and its BIC margin over the runner-up. cub denotes cubic, quad denotes quadratic, and lin denotes linear. $\sim$ marks a margin below 2, which is no meaningful preference.}
\label{tab:third_ray_form_selection}
\end{table}
\FloatBarrier

	\begin{figure}[h!]
		\centering
		\includegraphics{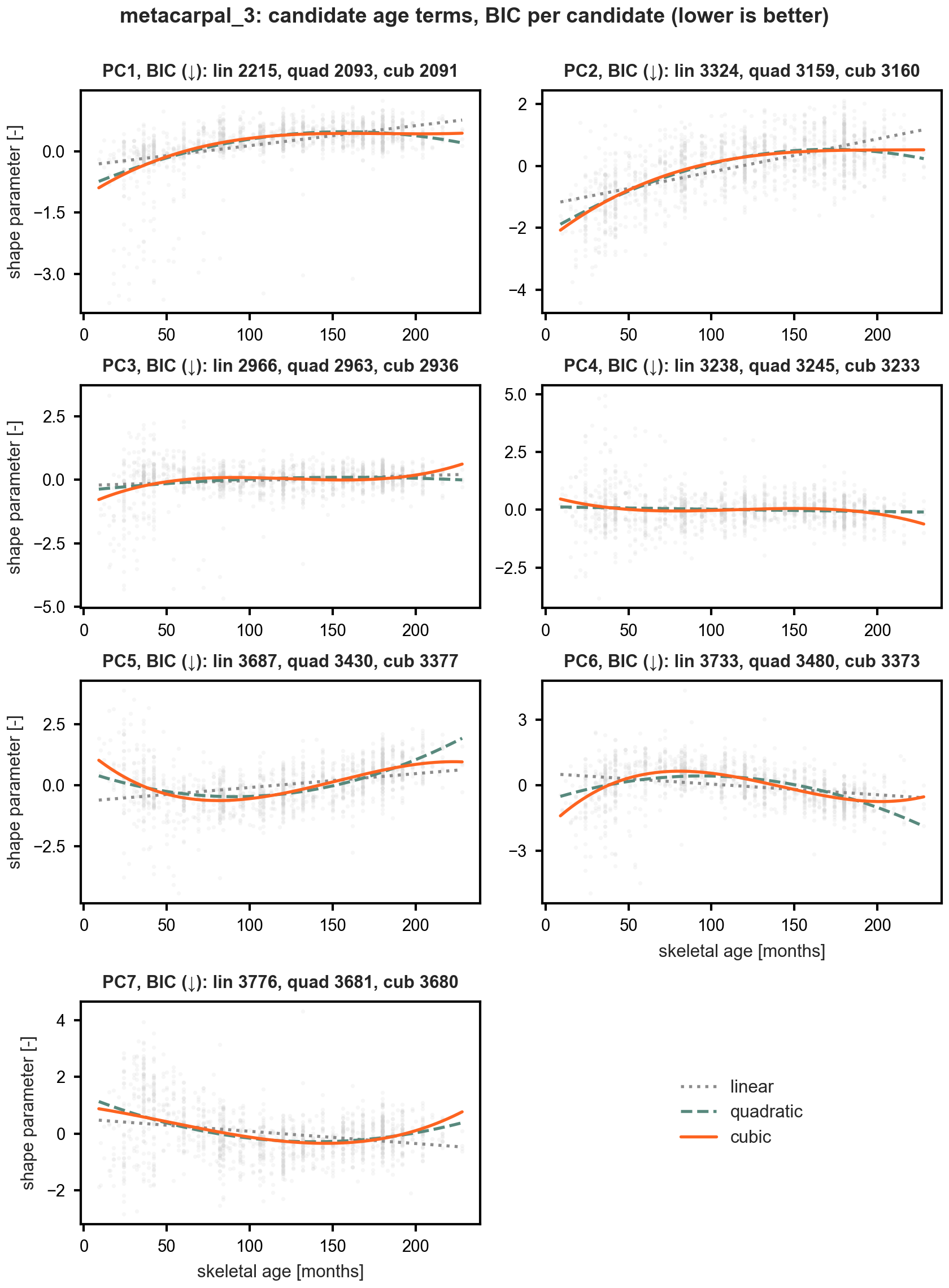}
		\caption{Comparison of the different fits of varying forms for the age terms. Each subfigure title denotes the difference in the Bayesian information criterion (BIC); a lower BIC indicates a better fit.}
	\end{figure}

\FloatBarrier

\section{Results from remaining bones}

\begin{table}[h!]
\centering
\resizebox{\ifdim\width>\textwidth \textwidth \else \width \fi}{!}{%
\begin{tabular}{ll|lllll}
\hline
Bone & modes & $T^2$ & $F$ & df & $D$ & $p$ \\
\hline
Distal phalanx 1 & PC1--8 & 61.712 & 7.677 & 8, 1447 & 0.4118 & $p \leq 0.001$ \\
Distal phalanx 2 & PC1--7 & 65.67 & 9.343 & 7, 1448 & 0.4248 & $p \leq 0.001$ \\
Distal phalanx 4 & PC1--7 & 51.661 & 7.35 & 7, 1448 & 0.3767 & $p \leq 0.001$ \\
Distal phalanx 5 & PC1--7 & 69.628 & 9.906 & 7, 1448 & 0.4374 & $p \leq 0.001$ \\
Medial phalanx 1 & PC1--7 & 108.65 & 15.46 & 7, 1448 & 0.5464 & $p \leq 0.001$ \\
Medial phalanx 3 & PC1--7 & 97.191 & 13.83 & 7, 1448 & 0.5167 & $p \leq 0.001$ \\
Medial phalanx 4 & PC1--6 & 46.595 & 7.739 & 6, 1449 & 0.3578 & $p \leq 0.001$ \\
Metacarpal 1 & PC1--7 & 189.57 & 26.97 & 7, 1448 & 0.7217 & $p \leq 0.001$ \\
Metacarpal 2 & PC1--6 & 121.68 & 20.21 & 6, 1449 & 0.5782 & $p \leq 0.001$ \\
Metacarpal 4 & PC1--9 & 80.775 & 8.926 & 9, 1446 & 0.4711 & $p \leq 0.001$ \\
Metacarpal 5 & PC1--9 & 62.104 & 6.862 & 9, 1446 & 0.4131 & $p \leq 0.001$ \\
Proximal phalanx 1 & PC1--7 & 153.42 & 21.83 & 7, 1448 & 0.6492 & $p \leq 0.001$ \\
Proximal phalanx 2 & PC1--7 & 128.8 & 18.32 & 7, 1448 & 0.5949 & $p \leq 0.001$ \\
Proximal phalanx 4 & PC1--8 & 99.923 & 12.43 & 8, 1447 & 0.5239 & $p \leq 0.001$ \\
Proximal phalanx 5 & PC1--9 & 98.077 & 10.84 & 9, 1446 & 0.5191 & $p \leq 0.001$ \\
\hline
\end{tabular}
}
\caption{Hotelling's $T^2$ over sex-stratified shape parameters encompassing $c$ shape modes retained for analysis. $F$ and $p$-values denote statistical significance whilst $T^2$ and $\mathcal{D}$ quantifies the distance between males and females. Results indicate that all bones are sexually dimorphic.}
\label{tab:dimorphism_omnibus_appendix}
\end{table}
\FloatBarrier

\begin{table}[htbp]
\centering
\resizebox{\ifdim\width>\textwidth \textwidth \else \width \fi}{!}{%
\begin{tabular}{l|lllllll}
\hline
Bone & Pillai & $s$ & Pillai$/s$ & $F$ & df & $p$ ($F$) & $p$ (Freedman--Lane) \\
\hline
Distal phalange 1 & 0.07335 & 3 & 0.02445 & 4.521 & 24, 4329 & $p \leq 0.001$ & $p \leq 0.001$ \\
Distal phalange 2 & 0.07228 & 2 & 0.03614 & 7.74 & 14, 2890 & $p \leq 0.001$ & $p \leq 0.001$ \\
Distal phalange 4 & 0.1063 & 3 & 0.03542 & 7.576 & 21, 4332 & $p \leq 0.001$ & $p \leq 0.001$ \\
Distal phalange 5 & 0.1161 & 3 & 0.03869 & 8.302 & 21, 4332 & $p \leq 0.001$ & $p \leq 0.001$ \\
Medial phalange 1 & 0.1503 & 3 & 0.05008 & 10.88 & 21, 4332 & $p \leq 0.001$ & $p \leq 0.001$ \\
Medial phalange 3 & 0.2033 & 3 & 0.06777 & 15 & 21, 4332 & $p \leq 0.001$ & $p \leq 0.001$ \\
Medial phalange 4 & 0.1335 & 3 & 0.04451 & 11.22 & 18, 4335 & $p \leq 0.001$ & $p \leq 0.001$ \\
Metacarpal 1 & 0.1125 & 3 & 0.0375 & 8.038 & 21, 4332 & $p \leq 0.001$ & $p \leq 0.001$ \\
Metacarpal 2 & 0.1001 & 3 & 0.03337 & 8.315 & 18, 4335 & $p \leq 0.001$ & $p \leq 0.001$ \\
Metacarpal 4 & 0.1789 & 3 & 0.05963 & 10.16 & 27, 4326 & $p \leq 0.001$ & $p \leq 0.001$ \\
Metacarpal 5 & 0.1652 & 3 & 0.05506 & 9.337 & 27, 4326 & $p \leq 0.001$ & $p \leq 0.001$ \\
Proximal phalange 1 & 0.148 & 3 & 0.04932 & 10.7 & 21, 4332 & $p \leq 0.001$ & $p \leq 0.001$ \\
Proximal phalange 2 & 0.1338 & 3 & 0.0446 & 9.629 & 21, 4332 & $p \leq 0.001$ & $p \leq 0.001$ \\
Proximal phalange 4 & 0.1593 & 3 & 0.05309 & 10.11 & 24, 4329 & $p \leq 0.001$ & $p \leq 0.001$ \\
Proximal phalange 5 & 0.1857 & 3 & 0.06189 & 10.57 & 27, 4326 & $p \leq 0.001$ & $p \leq 0.001$ \\
\hline
\end{tabular}
}
\caption{Results from Pillai's trace on sex $\times$ age interaction. Raw Pillai's trace magnitudes are normalized with respect to $s$ degrees of freedom, giving Pillai$/s$. We report raw $p$-values based on analytic $F$, but because residuals are heteroskedastic, we used permutation tests via Freedman-Lane sampling with Rademacher sign flipping to obtain a more robust indicator of significance. Results indicate that all bones have sex-specific ontogeny. }
\label{tab:interaction_omnibus_appendix}
\end{table}
\FloatBarrier

\vfill